# Connecting Humanities and Social Sciences: Applying Language and Speech Technology to Online Panel Surveys

**Henk van den Heuvel, Martijn Bentum, Simone Wills, Judith C. Koops**

CLS/CLST, Radboud University, Nijmegen, The Netherlands

Netherlands Interdisciplinary Demographic Institute (NIDI)-KNAW/University of Groningen, The Hague, The Netherlands

{henk.vandenheuvel, martijn.bentum, simone.wills}@ru.nl; koops@nidi.nl

**Abstract**

In this paper, we explore the application of language and speech technology to open-ended questions in a Dutch panel survey. In an experimental wave respondents could choose to answer open questions via speech or keyboard. Automatic speech recognition (ASR) was used to process spoken responses. We evaluated answers from these input modalities to investigate differences between spoken and typed answers. We report the errors the ASR system produces and investigate the impact of these errors on downstream analyses. Open-ended questions give more freedom to answer for respondents, but entail a non-trivial amount of work to analyse. We evaluated the feasibility of using transformer-based models (e.g. BERT) to apply sentiment analysis and topic modelling on the answers of open questions. A big advantage of transformer-based models is that they are trained on a large amount of language materials and do not necessarily need training on the target materials. This is especially advantageous for survey data, which does not contain a lot of text materials. We tested the quality of automatic sentiment analysis by comparing automatic labeling with three human raters and tested the robustness of topic modelling by comparing the generated models based on automatic and manually transcribed spoken answers.

## 1. Introduction

This paper reports on work carried out in *Task 4.4. Voice recorded interviews and audio analysis* in the Social Sciences and Humanities Open Cloud project (SSHOC[1]).

The SSHOC project is part of the INFRAEOSC 04-2018 activities to support actions outlined in the 2016 European Cloud Initiative Communication[2], to further integrate and consolidate e-infrastructure platforms, to connect ESFRI (European Strategy Forum on Research Infrastructures) infrastructures to the EOSC (European Open Science Cloud)[3], and to develop a European Data Infrastructure (EDI). In this sense, SSHOC contributes the social sciences and humanities, as well as heritage science, to the EOSC's overarching policy perspective.

SSHOC is not the only infrastructure bridging social sciences and the humanities. The need for interdisciplinary collaboration in exchanging and mutually exploiting data and tools has been expressed for over a decade (see e.g. Dușa et al, 2014; Samara and Scott, 2014) and is now vibrant both in national and international contexts. On a national level we mention the PDI-SSH initiative in the Netherlands[4] and the Berlin University Alliance in Germany[5].

In Task 4.4 SSHOC aims to integrate the collection of Computer Assisted Recorded Interviews (CARI) into an existing social sciences survey infrastructure. In this task project representatives of the Netherlands Interdisciplinary Demographic Institute (Social Sciences), and CLARIN-ERIC (Humanities) work closely together.

Every year, survey infrastructures such as the Generations and Gender Programme (GGP), the European Values Study (EVS), the European Social Survey (ESS), and the Survey of Health, Ageing, and Retirement in Europe (SHARE) conduct tens of thousands of interviews across Europe. People are chosen at random from the population and queried about a variety of topics that are useful to researchers and policymakers. However, much of the information communicated during an interview is lost. The tone, clarity, fluidity, and vocabulary depth of the replies can all be used to gain insight into many aspects of interest to social scientists, such as cognitive function and verbal reasoning skills. To make use of this lost data, language and speech technology (LST) are of significant importance when integrated into the survey infrastructures' analytical pipeline.

In Emery et al. (2019) guidelines for the integration of LST into CARI are set out. The study reported in the present paper expands on some first analyses reported in Koops et al. (2021). In this paper we explore the application of automatic speech recognition (ASR) to audio recordings of participant responses, and perform follow-up text analyses on the resulting texts: sentiment analysis and topic modelling. In particular, we compare the results obtained from the audio input with the keyboard input, as well as the results obtained from the raw ASR output with manually corrected ASR transcripts. These comparisons will tell us more about the added value of A. the modality of audio responses, and B. the required quality of transcriptions of audio responses for follow-up analyses.

---

[1] https://sshopencloud.eu/
[2] https://www.eumonitor.eu/9353000/1/j9vvik7m1c3gyxp/vk3g67wp8uw7
[3] https://eosc-portal.eu/
[4] https://pdi-ssh.nl/en/
[5] https://www.myscience.org/news/wire/research_infrastructures_in_the_humanities_and_social_sciences-2021-FUB

## 2. Data and Method

### 2.1 Data

The collection of voice recordings was embedded in the Longitudinal Internet studies for the Social Sciences (LISS)[6]. LISS is an online panel survey. The LISS panel consists of 5,000 Dutch households with around 7,500 people aged 16 and older. The panel is updated annually and is based on a genuine random sample of households collected from Statistics Netherlands' population record. Every month, panel members take roughly 30 minutes to answer online questions. Panelists receive a monetary incentive of €2.50 for completing a 10-minute survey. The LISS Core Study receives a portion of the interview time available in the LISS panel. The LISS Core Study is conducted once a year and is aimed to track changes in the panel members' lives and living conditions. There is plenty of room to collect data for other scientific, policy, or socially relevant research in addition to the LISS Core Study.

In April 2021, the Audio Module was integrated into the LISS panel. A total of 771 LISS panelists, ranging in age from 20 to 49, were asked to participate. The interviewing duration for the Audio Module was around 10 minutes.

The Audio Module began with inquiries regarding the respondent's device, internet browser, and microphone availability. The ability to record one's voice was determined based on the answers to these questions. If respondents were able to record, they were directed to the Audio Module, where they were requested to consent to the recording of their microphone for the duration of the Audio Module. The microphone was tested once before the module began to verify its quality. Respondents whose device couldn't make a voice recording or who didn't allow permission to use the microphone were given the option of closing the questionnaire and continuing at a later time or continuing with a written version of the Audio Module. Respondents who chose the latter were presented the same questions as those who chose the former, but they were required to submit their answers in a text box.

For ASR purposes the Questfox speech recognizer[7] was integrated into the LISS panel software. After the data was collected, the LISS team carefully reviewed all voice recordings to ensure that all personally identifying information, such as names and addresses, was removed from the audio files. Concretely, this meant that the LISS team listened to all audio files and deleted personal information manually. Speech-to-text transcripts of the audio recordings were also provided by the Questfox software. The LISS team made sure that any personally identifiable information was removed from these transcripts as well.

In total, 771 people between the ages of 20 and 49 were invited to participate in the Audio Module. 631 participants (82 percent) started the module, and 486 (63 percent) finished it. 100 (21%) of individuals who finished the module used voice recordings to respond, whereas 386 (79%) typed their replies in a text box. This raises the total response rate for audio recordings to 13%, and the response rate for written comments to 50%. This resulted in 2379 audio files from 760 sessions (some respondents had several recording sessions), with a total duration of nearly 9 hours of audio.

The Questfox recognition output was available for 2165 recordings. After processing of the recordings and transcriptions, which included matching the transcriptions to their respective audio recording, a final set of 1796 recordings and matching transcriptions, totaling 7 hours 15 minutes, were used for the analyses presented in the paper.

The Audio Module consists of 15 core, open-ended questions, covering topics of *democracy*, *the European Union, trust* and *marriage*. These were selected for audio recording because of their potential to evoke longer and more emotional responses. These questions are supplemented by 7 Likert scale questions establishing baseline opinions, and 5 closing questions regarding the survey experience. Only the core 15 questions and one question for testing audio recording allowed for audio responses.

The questions and their associated IDs can be found in the Appendix together with their English translations. For the various analyses reported below different subsets of the questions were used. We will indicate the questions involved in each of the substudies by their question IDs.

### 2.2 Perceptual Evaluation

High quality audio recordings are pivotal for the use of ASR tools to prepare speech-to-text transcripts (Draxler et al., 2020). They determine how well the collected information can be used for further data exploration.

Project partner Centre for Language and Speech Technology (CLST) performed various analyses of the audio quality of the recordings. The first analysis was of a qualitative nature and was carried out by listening to a subset of the recordings. The selection comprised 594 audio files of all available answers to questions Q8, Q13, Q14, Q42, Q43. Most of the subset of 150 speakers were thus included at least three times in the selection. Each recording could obtain one of the following labels: *Good, Average, Poor, Very Poor*. Comments were added if utterances were considered *Poor* or *Very Poor*.

### 2.3 Quality of ASR transcriptions

The accuracy of the ASR transcriptions is pivotal for the reliability of any further data analytics carried out on the participant responses. To assess the quality of the Questfox transcriptions the Word Error Rate (WER) was calculated, with manually-corrected ASR transcriptions serving as the golden-standard.

---

[6] LISS Panel: https://www.lissdata.nl/

[7] https://www.questfox.com/en/

The reference transcripts were made by manually correcting the ASR output of the Questfox recognizer. This is a significantly faster and less labour intensive approach compared to transcribing audio from scratch. Although, it is important to note that this approach does introduce an element of bias towards the recognition output being used as the starting point.

In addition to scoring the Questfox ASR output, the recognition results of three CLST speech recognition engines[8] were also scored, to serve as a point of comparison. These three speech recognition engines are respectively tuned for daily conversations (DC), oral history interviews (OH), and parliamentary debates (PD).

The NIST Sclite[9] toolkit was used to calculate the WER scores, using text-normalised versions of the transcriptions.

## 2.4 Sentiment Analysis

To assess the performance of automatic sentiment analysis on questionnaire materials, we selected questions that might prompt positive or negative sentiment in the answers. For the speech response condition we selected question 13, 15, 20, 22 and 29. For the keyboard response condition we selected the equivalent questions 16, 18, 23, 25 and 32. For each of these questions we selected all available answers (N = 2046), except the answers equal to: 'ik weet (het) niet' *I don't know.*

Three human raters rated each answer as either *positive*, *negative* and *neutral*. The neutral label was used to filter out answers and questions lacking positive or negative sentiment. Based on the human sentiment ratings we excluded questions 13 and 16 (this is the same question for the audio and keyboard input condition respectively), because 73% of the corresponding answers were rated neutral. In addition, we excluded each answer that one of the human raters rated as neutral. This resulted in a set of 744 answers used for the comparison between human and automatic sentiment raters. We used the Dutch transformer model BERTje[10] (de Vries et al., 2019) fine-tuned on the Dutch Book Reviews Dataset (van der Burgh and Verberne, 2019) for automatic sentiment analysis. The automatic sentiment analysis provided either a positive or a negative label.

The Fleiss' Kappa metric was used to measure agreement between all raters. In addition to the Fleiss' Kappa metric, we created a ground truth based on a majority-rule of the three human ratings for each answer and used the ground truth together with the automatic sentiment ratings to compute an F1 score.

## 2.5 Topic Modelling

To assess whether the ASR transcriptions adversely influence topic modelling of the answers compared to human transcriptions, we created a *manual* and an *automatic* transcription dataset. For these datasets we selected all available answers (N = 5987) corresponding to the following questions: 13, 14, 15, 20, 21, 22, 27, 28, 29, 34, 35, 36, 41, 42 and 43.

We included both materials from the keyboard and speech input condition to have enough text materials for topic modelling. The crucial difference between the *manual* and *automatic* dataset is that for the speech input condition, we either utilized the answer as transcribed by the manually transcriber or as transcribed by the Questfox ASR system. Text materials from the keyboard input condition are the same for both datasets.

For topic modelling, we used a method proposed by Grootendorts[11], which utilizes a sentence-transformer model in combination with clustering and TF-IDF to generate topics. The first step is to map each answer to a 512 dimensional vector with the aid of multilingual Bert model[12] (Reimers & Gurevych, 2019).

The vectors from step 1 are clustered with the aid of Hierarchical Density-Based Spatial Clustering of Applications with Noise (HDBSCAN) as implemented in Python[13] (see also McInnes & Healy, 2017). The advantage of this clustering algorithm is that the number of clusters does not need to be specified, only the minimum cluster size. In addition, it allows for outliers, which are not forced to be part of a cluster.

Subsequently we applied TF-IDF on the clustered answers, whereby the answer-clusters were used as the documents. This resulted in a value for each word for each cluster that indicated the importance of a given word for a given cluster. To create a topic representation we extracted the top 20 words per topic based on their TF-IDF scores.

To aid with the selection of the minimum cluster size, we utilized the topic coherence metric as implemented in the Python package Gensim[14], see also (Röder, Both & Hinneburg, 2015). We selected the top 20 words for each topic and utilized the 'u_mass' coherence metric, which is based on the smoothed conditional probability of each top word pair (Röder, Both & Hinneburg, 2015). We ran the clustering algorithm with minimum cluster sizes ranging from 2 - 50, applied the TF-IDF as discussed before and selected the cluster size that resulted in the best coherence score with the added restriction that there should be more than one topic.

The selected questions were grouped into four sets, relating to *democracy* (questions 13, 14, 15, 16, 17, 18),

---

[8] https://webservices.cls.ru.nl/oralhistory
[9] https://www.nist.gov/itl/iad/mig/tools .
[10] https://huggingface.co/wietsedv/bert-base-dutch-cased-finetuned-sentiment
[11] https://towardsdatascience.com/topic-modeling-with-bert-779f7db187e6
[12] https://huggingface.co/sentence-transformers/distiluse-base-multilingual-cased-v2
[13] https://github.com/scikit-learn-contrib/hdbscan
[14] https://radimrehurek.com/gensim/models/coherencemodel.html

*Europe* (questions 20, 21, 22, 23, 24, 25), *trust* (questions 27, 28, 29, 30, 31, 32) and *marriage* (questions 34, 35, 36, 37, 38, 39, 41, 42, 43, 44, 45, 46).

For each set we applied the topic modelling method described before for both the automatic transcriptions and the manually corrected transcriptions. Subsequently we compared the topics generated for the automatic and manual transcriptions by computing Spearman's rank correlation coefficient for the top 100 words of each topic.

## 3. Results and Discussion

### 3.1 Speech and text input comparison

Respondents were free to choose between speech or keyboard input for filling out the questionnaire. We compared the answers on length and distribution of word types, see Table 1. The data suggests that respondents typically provide longer answers when using speech compared to keyboard. The modalities do not appear to influence the percentage of content words.

|  | Speech | Keyboard |
|---|---|---|
| # responses | 1,665 | 4,322 |
| median # words | 16 | 9 |
| average # words | 25.96 | 12.09 |
| max # words | 139 | 209 |
| total # words | 43,216 | 52,249 |
| median # content words | 13 | 6 |
| average # content words | 18.9 | 8.55 |
| total # content words | 30,539 | 36,974 |
| percentage content words | 70.69% | 70.76% |

Table 1: Comparison of speech and keyboard input modality for questionnaire answers.

### 3.2 Perceptual Evaluation

Table 2 displays the results of the perceptual evaluation carried out by listening to a subset of the recordings and assessing their acoustic quality.

| Label | Frequency | Percentage |
|---|---|---|
| Good | 338 | 56.90% |
| Average | 187 | 31.48% |
| Poor | 53 | 8.92% |
| Very poor | 16 | 2.69% |

Table 2: Perceptual assessment of the audio recordings

About 90% of the recordings are of a good or average acoustic quality. These are well suited for ASR. The remaining recordings showed flaws in terms of loud background noise by other speakers or equipment, clipping, channel flaws, and inadequate microphone settings. ASR is also hampered if there is not enough silence at the start of a recording. In some (but fewer) cases the problematic quality is due to characteristics of the speaker, such as strong accents, fast or hesitant articulation, or a very low or loud speaking volume.

### 3.3 Quality of ASR transcriptions

The performance of the four speech recognisers, in terms of WER, is presented in Table 3. The Questfox ASR is seen to perform the best, outperforming the other engines by around 10 - 12 %. This is partly due to the bias in manual transcriptions (see 2.3) and to the fact that this recogniser was better trained towards the target material.

| Label | WER | subs | del | ins |
|---|---|---|---|---|
| Questfox | 24.7 | 9.19 | 13.97 | 1.54 |
| DC | 34.34 | 14.51 | 17.12 | 2.71 |
| OH | 36.51 | 15.54 | 18.23 | 2.73 |
| PD | 34.26 | 14.48 | 17.07 | 2.71 |

Table 3: Performance in Word Error Rate (WER) for the various speech recognisers.

For all systems, deletions contributed the most to the WER, followed by substitutions. Looking closer at the errors produced by Questfox, as the best performing system, several trends can be noted.

In terms of deletions, the most frequently deleted words are monosyllabic words. This is reflected in the 25 most frequent deletions, responsible for 42.89% of deletions, all of which are monosyllabic. These words pose a challenge to acoustic modelling due to the limited acoustic information they provide.

25.55% of the deleted words are not present in any of the text produced by the speech recogniser. For the substituted reference words, this is even higher at 34.18%. This accounts for 5% and 10.98% of all instances of deletion and substitution, respectively.

This demonstrates that domain-tuning of the ASR to the topics of a panel survey would be an important aspect for improving recognition performance.

### 3.4 Sentiment Analysis

The Fleiss' Kappa metric ranges from -1 to 1, corresponding to perfect disagreement to perfect agreement. Fleiss' Kappa was computed for the three human raters, judging the sentiment of answers $\kappa = .96$. The raters agreed perfectly 725 times out of a possible 744. Additionally, we computed the Fleiss' Kappa for the human raters and the automatic sentiment labels with fine-tuned BERTje for the same answer set $\kappa = .73$. The raters agreed perfectly 549 times out of 744. Lastly, we created a ground truth based on a majority-rule of the three human annotators and computed the precision, recall and F1-score of the automatic sentiment analysis, see Table 4. The automatic sentiment analysis shows a bias towards negative labels compared to the human raters.

| Label | Precision | Recall | F1 |
|---|---|---|---|
| Negative | 0.61 | 0.93 | 0.73 |
| Positive | 0.94 | 0.66 | 0.78 |

Table 4: Performance results of the sentiment analysis with fine-tuned Bertje, ground truth based on majority rule of three human raters.

With the development of transformer-based pre-trained language models such as BERT it is possible to achieve good performance without training a dataset specific model. We applied a Dutch transformer model Bertje fine-tuned for sentiment analysis (de Vries et al., 2019) to test the performance on a dataset from a different domain. The model was fine-tuned on a Dutch book review dataset (van der Burgh and Verberne, 2019) and applied on a subset of our questionnaire materials.

The questionnaire materials were rated by three human raters with the labels positive, negative and neutral. The neutral label was incorporated to exclude answers without sentiment. For the remaining materials the agreement between human and automatic raters was assessed with Fleiss' Kappa and an F1-score.

The human raters showed high agreement on the non-neutral materials. The Fleiss' Kappa in which the automatic sentiment analysis was included showed lower agreement. The precision and recall metric revealed that the automatic sentiment analysis has a negative label bias compared to the human raters. This could be due to a difference between the materials the model was fine-tuned and applied on.

One challenge with fully automating sentiment analysis using the current approach are the neutral responses. We filtered the neutral responses, based on human labelling. The neutral label can be subdivided into two groups. Neutral because the positive and negative sentiment is balanced, or neutral because the answer does not relate to the question. Many of the neutral answers were variants of the none-response *I don't know*. This could be a good start of filtering out materials that cannot be sensibly rated on sentiment. Furthermore, it is important to screen questions to assess whether one would expect an answer with positive or negative sentiment. For example, the question we excluded based on the many neutral ratings: *What are the most important characteristics of a democracy according to you?* Was in hindsight not a question that prompts a sentiment laden answer. Lastly, extending the automatic sentiment classifier with a neutral label, potentially subdividing balanced sentiment and non-answers would improve usability in a fully automated sentiment analysis setup.

### 3.5 Topic Modelling

We compared the topics of the *manual* and *automatic* datasets. Topics were generated for four answer sets related to questions about *democracy, Europe, trust* and *marriage*. The comparison was done by computing Spearman's rank correlation coefficient for the top 100 words between topics in the manual and automatic datasets for each answer set. We set a conservative threshold at a correlation of .7 and p value < .05 to consider topics for the manual and automatic datasets as similar. In Table 5 the number of topics similar for automatic and manual datasets are listed. We also computed the percentage of texts in the manual and automatic datasets that would be clustered similarly based on the generated topics.

| Answer set | # topics manual | # topics automatic | # topics similar | % texts in similar cluster |
|---|---|---|---|---|
| Democracy | 19 | 50 | 8 | 43% |
| Europe | 15 | 15 | 8 | 66% |
| Trust | 6 | 4 | 4 | 75% |
| Marriage | 13 | 10 | 8 | 64% |

Table 5: Number of topics for manual and automatic datasets and the number of similar topics; percentage of speech input texts similarly clustered.

The results listed in Table 5 indicate that the difference between automatic and manual transcriptions influences the generated topics. The number of topics generated for automatic and manual transcriptions differs quite extensively for the *democracy* answer set, while for *Europe*, *trust* and *marriage* the number of topics is identical or more similar.

The column reporting the percentage of speech input texts that are in a similar cluster indicates that for text clustering purposes the algorithm attains somewhat more similar outcomes for automatic and manual texts. This could be because not every topic is related to an equal number of texts. The topics related to more texts appear to be more robust against the noise introduced by the automatic transcriptions.

The representation of topics is influenced by the difference between automatic and manual transcriptions. For example, Tables 6 and 7 give examples of similar and dissimilar topics found for the manual and automatic transcriptions.

| Transcription | top 5 words (Spearman's R = .96) |
|---|---|
| Manual | iedereen, mening, mag, mee, telt 'everyone, opinion, may, with, counts' |
| Automatic | iedereen, mening, gelijk, mag, telt 'everyone, opinion, equal, may, counts' |

Table 6: Example of a highly similar topic between manual and automatic transcriptions. Spearman's R is based on the ordering of the top 100 words.

| Transcription | top 5 words (Spearman's R = .44) |
|---|---|
| Manual | leven, maakt, mekaar leiden, land 'life, make, each other, lead, country' |
| Automatic | zeggen, samenleving, mens, iedereen, ieder 'say, society, human, everyone, each' |

Table 7: Example of a dissimilar topic between manual and automatic transcriptions. Spearman's R is based on the ordering of the top 100 words.

The results indicate that topic modelling based on automatic transcriptions will generate some highly similar topics compared to those based on manual transcriptions. However, many topics will differ as well. A better ASR performance will obviously bring both results closer together.

Another source of variability might be the methodology used to generate the topics. The number of clusters found based on the word embeddings differed quite extensively for the *democracy* set. This could indicate that the clustering technique is overly sensitive to the noise generated by the word errors of the ASR output. Alternatively, the embeddings generated with the transformer based multilingual BERT model might be very sensitive to transcription errors in the ASR output.

## 4. Conclusion

The collection of a CARI module was successfully integrated into LISS, a Dutch social sciences panel survey, in Task 4.4. of the SSHOC project. In April 2020, respondents aged 20-49 completed the Audio Module, which consisted of a mix of closed and open-ended questions.

Audio recordings are a valuable source of information. Acoustic data can be used to obtain additional information about the respondents, such as their emotional state, verbal reasoning skills, language proficiency, and dialect, in addition to the factual data supplied. These studies necessitate high-quality recordings. Because respondents are already near to the device and its microphone while filling out the questionnaire and there is little to no speaker overlap, the LISS panel and other social surveys could be good candidates for producing high-quality audio data. The experiment corroborated this. 90% of the audio recordings gathered by the LISS panel were acceptable for further analysis.

Linking data from higher-order linguistic and computational analysis with other survey content can add significant value to current social science infrastructures. It's also true in the other direction. In the analysis and interpretation of enormous amounts of (online) digital data, ASR and other linguistic and computational methods are critical. However, the development and implementation of these tools is skewed, and they tend to work less optimally for the majority of languages. Even though 90% of the Dutch recordings were of sufficient quality for ASR, the Word Error Rate was between 25-37%, indicating that there is ample room for improvement of the ASR engines.

Quantitative analytics, such as topic modelling and sentiment analysis, can be performed on the data, as we illustrated in this paper. However, improvements such as a method to filter out neutral responses for sentiment analysis and improvements to the WER of automatic transcriptions for topic modelling on spoken answers are essential for a fully automatic pipeline and high quality output.

A digital data collection in multiple nations might provide a similar test for other languages and could be used to create an inventory of words and expressions that is needed to improve ASR performance. Furthermore, the rich socioeconomic and demographic background data in such surveys would allow for a full study of which groups of speakers the ASR technologies operate best for and which groups need to be improved.

**Acknowledgements**

SSHOC, "Social Sciences and Humanities Open Cloud", has received funding from the European Union's Horizon 2020 project call H2020-INFRAEOSC-04-2018, grant agreement #823782

# Language Resource References

The speech-to-text transcripts devoid of identifiable information, as well as the responses to the survey questions will be made available in excel format to researchers via the LISS Data Archive[15] from January 2022 onward. The data will be published in the LISS panel data archive[16] as an "Assembled Study" and can be found under the title "Voice recorded interviews SSHOC project".

---

[15] Data https://www.dataarchive.lissdata.nl/
[16] https://www.dataarchive.lissdata.nl/

# Appendix: Questions Used in the Study

Overview of questions used in the study with their English translations.

| Question ID (Audio) | Question ID (Keyboard) | Dutch text | English translation |
|---|---|---|---|
| Q08 (Test) | | De wereld is mijn speeltuin | The world is my playground |
| Q13 | Q16 | Wat zijn voor u de belangrijkste kenmerken van een democratie? | What do you consider to be the most important features of a democracy? |
| Q14 | Q17 | Waarom zijn deze kenmerken zo belangrijk? | Why are these characteristics so important? |
| Q15 | Q18 | Hebt u het gevoel dat Nederland op dit moment democratisch wordt bestuurd? Waarom wel of waarom niet? | Do you feel that the Netherlands is currently governed democratically? Why or why not? |
| Q20 | Q23 | Wat betekent het voor u om een Europeaan of een burger van de Europese Unie te zijn? | What does it mean for you to be a European or a citizen of the European Union? |
| Q21 | Q24 | Wat zijn volgens u kenmerken van een gedeelde Europese cultuur? | What do you think are the characteristics of a shared European culture? |
| Q22 | Q25 | Voelt het voor u alsof u deel uitmaakt van de Europese Unie? Waarom wel of waarom niet? | Do you feel as if you are part of the European Union? Why or why not? |
| Q27 | Q30 | Wat zijn voor u de belangrijkste redenen om een persoon te kunnen vertrouwen? | What do you consider to be important reasons to trust a person? |
| Q28 | Q31 | Welke redenen zou u kunnen bedenken om iemand niet te vertrouwen? | What reasons can you think of for not trusting someone? |
| Q29 | Q32 | Een samenleving waarin u andere mensen kunt vertrouwen, maakt het dagelijks leven gemakkelijker. Hebt u het gevoel dat u in een betrouwbare samenleving leeft? En waarom hebt u dit gevoel? | A society in which you can trust other people makes living easier. Do you feel you life in a trustworthy society? Why do you feel this way? |
| Q34 | Q37 | Sommige stellen zijn getrouwd, terwijl andere stellen met elkaar samenwonen zonder te trouwen. Wat zijn volgens u de voordelen van trouwen? | Some couples are married while other couples live together without getting married. What do you consider to be the advantages of marriage? |
| Q35 | Q38 | Wat zijn volgens u de nadelen van trouwen? | What do you think are the disadvantages of getting married? |
| Q36 | Q39 | Waarom denkt u dat mensen ervoor kiezen om te gaan trouwen? | Why do you think people choose to get married? |
| Q41 | Q44 | Wat zijn volgens u de voordelen van ongehuwd samenwonen? | What do you think are the advantages of unmarried cohabitation? |
| Q42 | Q45 | Wat zijn volgens u de nadelen van ongehuwd samenwonen? | What do you think are the disadvantages of unmarried cohabitation? |
| Q43 | Q46 | Waarom denkt u dat stellen die al een tijdje samenzijn besluiten niet te trouwen? | Why do you think couples who have been together for a while decide not to get married? |